\documentclass[twoside]{article}
\usepackage{ecj,palatino,latexsym,natbib}
\usepackage{booktabs}
\usepackage{graphicx}
\graphicspath{{images/}}
\usepackage{subfigure}
\usepackage{url}
\usepackage{hyperref}
\hypersetup{
        hidelinks,
	colorlinks=true,
	allcolors=black,
	pdfstartview=Fit,
	breaklinks=true
}
\usepackage{algorithm}
\usepackage{algpseudocode}
\usepackage{amsmath}

\DeclareUnicodeCharacter{2212}{-}
\begin{document}

\ecjHeader{x}{x}{xxx-xxx}{200X}{45-character paper description goes here}{D. YANG ET AL.}
\title{\bf An Improved Genetic Algorithm and Its Application in Neural Network Adversarial Attack}  

\author{\name{\bf Dingming Yang} \hfill \addr{202071544@yangtzeu.edu.cn}\\ 
        \addr{School of Computer Science, Yangtze University, Jingzhou, 434023, China}
\AND
        \name{\bf Zeyu Yu} \hfill \addr{yuzeyu\_jz@163.com}\\
        \addr{School of Electronic \& Information, Yangtze University, Jingzhou, 434023, China}
        \name{\bf Hongqiang Yuan} \hfill \addr{429809060@qq.com}\\
        \addr{School of Urban Construction, Yangtze University, Jingzhou, 434000, China}\\
        \name{\bf Yanrong Cui} \hfill \addr{cyanr@yangtzeu.edu.cn}\\
        \addr{School of Computer Science, Yangtze University, Jingzhou, 434023, China}
}

\maketitle

\begin{abstract}
\figurename
\tablename
The choice of crossover and mutation strategies plays a crucial role in the searchability, convergence efficiency and precision of genetic algorithms.
In this paper, a novel improved genetic algorithm is proposed by improving the crossover and mutation operation of the simple genetic algorithm, and it is verified by 15 test functions.
The qualitative results show that, compared with three other mainstream swarm intelligence optimization algorithms, the algorithm can not only improve the global search ability, convergence efficiency and precision, but also increase the success rate of convergence to the optimal value under the same experimental conditions. 
The quantitative results show that the algorithm performs superiorly in 13 of the 15 tested functions.
The Wilcoxon rank-sum test was used for statistical evaluation, showing the significant advantage of the algorithm at $95\%$ confidence intervals.
Finally, the algorithm is applied to neural network adversarial attacks.
The applied results show that the method does not need the structure and parameter information inside the neural network model, and it can obtain the adversarial samples with high confidence in a brief time just by the classification and confidence information output from the neural network.

\end{abstract}


\section{Introduction}

In real life, optimization problems such as shortest path, path planning, task scheduling, parameter tuning, etc. are becoming more and more complex and have complex features such as nonlinear, multi-constrained, high-dimensional, and discontinuous \citep{deng2021improved}.
Although a series of artificial intelligence algorithms represented by deep learning can solve some optimization problems, they lack mathematical interpretability due to the existence of a large number of nonlinear functions and parameters inside their models, so they are difficult to be widely used in the field of information security.
Traditional optimization algorithms and artificial intelligence algorithms can hardly solve complex optimization problems with high dimensionality and nonlinearity in the field of information security.

Therefore, it is necessary to find an effective optimization algorithm to solve such problems.
In this background, various swarm intelligence optimization algorithms have been proposed one after another, such as Particle Swarm Optimization(PSO)\citep{kennedy1995particle,eberhart1995new},Grey Wolf Optimizer(GWO)\citep{mirjalili2014grey}, etc.
Subsequently, a variety of improved optimization algorithms also have been proposed one after another.
For example, the improved genetic algorithm for cloud environment task scheduling\citep{zhou2020improved}, the improved genetic algorithm for flexible job shop scheduling\citep{zhang2020improved}, the improved genetic algorithm for green fresh food logistics\citep{li2020optimization}, etc.

However, these improved optimization algorithms are improved for domain-specific optimization problems and do not improve the accuracy, convergence efficiency and generalization of the algorithms themselves.
In this paper, the crossover operator and mutation operator of the genetic algorithm are improved to improve the convergence efficiency and precision of the algorithm without affecting the effectiveness of the improved genetic algorithm on most optimization problems.
The effectiveness of the improved genetic algorithm is also verified through many comparison experiments and applications in the field of neural network adversarial attacks.

The main contributions of this paper are as follows:
\begin{itemize}
\item By improving the single-point crossover link of SGA, the fitness function is used as an evaluation index for selecting children after crossover, thus reducing the number of iterations and accelerating the convergence speed.
\item By improving the basic bitwise mutation of the SGA, traversing each gene of the offspring and performing selective mutation on them, setting different mutation rates for two parts of a chromosome, thus improving the global search in the stable case of local optimum.
\item The improved genetic algorithm is applied to the field of neural network adversarial attack, which increases the speed of adversarial sample generation and improves the robustness of the neural network model.
\end{itemize}

\section{Related Works}
\subsection{Genetic Algorithm}
\begin{figure}[ht]
        \centering
        \includegraphics[width=0.9\textwidth]{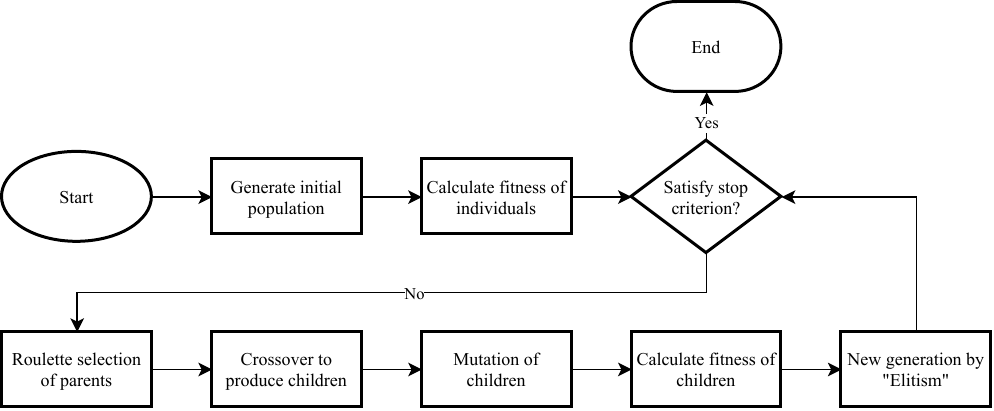}
        \caption{Genetic algorithm flowchart}
        \label{fig:ga_flowchart}
\end{figure}

Genetic Algorithm is a series of simulation evolutionary algorithms proposed by \cite{holland1975adaptation}, and later summarized by DeJong, Goldberg and others.
The general flowchart of the Genetic Algorithm is shown in \figurename~\ref{fig:ga_flowchart}.
The Genetic Algorithm first encodes the problem, then calculates the fitness, then selects the parent and the mother by roulette, and finally generates the children with high fitness by crossover and mutation, and finally generates the individuals with high fitness after many iterations, which is the satisfied solution or optimal solution of the problem.
Simple Genetic Algorithm (SGA) uses single-point crossover and simple mutation to embody information exchange between individuals and local search, and does not rely on gradient information, so SGA can find the global optimal solution.

\subsection{Other Meta-heuristic Algorithms}
The meta-heuristic algorithm is problem-independent, does not exploit the specificity of the problem, and is a general solution.
In general, it is not greedy, can explore more search space, and tends to obtain the global optimum.
To be more specific, meta-heuristic have one of the most important ideas: a dynamic balance mechanism between diversification and intensification.

The PSO\citep{kennedy1995particle,eberhart1995new} algorithm is a swarm intelligence-based global stochastic search algorithm inspired by the results of artificial life research and by simulating the migration and flocking behavior of bird flocks during foraging, and its basic idea is inspired by the results of research on modeling and simulation of birds flock behavior.
The GWO algorithm is a swarm intelligence optimization algorithm proposed by \cite{mirjalili2014grey}. The algorithm is inspired by the grey wolf prey hunting activity and developed as an optimization search algorithm, which has strong convergence performance, few parameters, and easy implementation.
The Marine Predator Algorithm (MPA)\citep{2020Marine} is mainly inspired by foraging strategies widely found in marine predators, namely Lévy and Brownian motion, and optimal encounter rate strategies in biological interactions between predators and prey.
The Artificial Gorilla Troops Optimizer (GTO)\citep{abdollahzadeh2021artificial}  was inspired by the gorilla group life behavior. The GTO is characterized by fast search speed and high solution accuracy.
The African Vulture Optimization Algorithm(AVOA)\citep{abdollahzadeh2021african} was inspired by the foraging and navigation behavior of African vultures. this algorithm is fast and has high solution accuracy which is widely used in single-objective optimization.
The Remora Optimization Algorithm (ROA)\citep{jia2021remora} first proposed an intelligent optimization algorithm inspired by the biological habits of the neutrals in nature, which has good solution accuracy and high engineering practical value in both function seeking to solve extreme values and typical engineering optimization problems.

\subsection{Neural Network Adversarial Attack}
\cite{szegedy2013intriguing} first demonstrated that a highly accurate deep neural network can be misled to make a misclassification by adding a slight perturbation to an image that is imperceptible to the human eye, and also found that the robustness of deep neural networks can be improved by adversarial training.
Such phenomena are far-reaching and have attracted many researchers in the area of adversarial attacks and deep learning security.
\cite{akhtar2018threat} surveyed 12 attack methods and 15 defense methods for neural networks adversarial attacks.
The main attack methods are finding the minimum loss function additive term \citep{szegedy2013intriguing}, increasing the loss function of the classifier \citep{kurakin2016adversarial}, the method of limiting the l\_0 norm \citep{papernot2016limitations}, changing only one pixel value \citep{su2019one}, etc.

\cite{nguyen2015deep} continued to explore the question of "what differences remain between computer and human vision" based on \cite{szegedy2013intriguing}.
They used the Evolutionary Algorithm to generate high-confidence adversarial images by iterating over direct-encoded images and CPPN (Compositional Pattern-Producing Network) encoded images, respectively.
They obtained high-confidence adversarial samples (fooling images) using the Evolutionary Algorithm on a LeNet model pre-trained on the MNIST dataset \citep{lecun1998mnist} and an AlexNet model pre-trained on the ILSVRC 2012 ImageNet dataset \citep{deng2009imagenet,russakovsky2015imagenet}, respectively.

Neural network adversarial attacks are divided into black-box attacks and white-box attacks. Black-box attacks do not require the internal structure and parameters of the neural network, and the adversarial samples can be generated with optimization algorithms as long as the output classification and confidence information is known.
The study of neural network adversarial attacks not only helps to understand the working principle of neural networks but also increases the robustness of neural networks by training with adversarial samples.

\section{Approaches}
This section improves the single-point crossover and simple mutation of SGA.
The fitness function is used as the evaluation index of the crossover link, and the crossover points of the whole chromosome are traversed to improve the efficiency of the search for the best.
A selective mutation is performed for each gene of the children's chromosome, and the mutation rate of the latter half of the chromosome is set to twice that of the first half to improve the global search under the stable situation of local optimum.
\newpage
\subsection{Improved Crossover Operation}
As shown in algorithm \ref{alg:crossover} is the Python pseudocode for the improved crossover algorithm.
The single-point crossover of SGA is to generate a random number within the parental chromosome length range, and then intercept the first half of the father's chromosome and the second half of the mother's chromosome to cross-breed the children according to the generated random number.
In this paper, the algorithm is improved by trying to cross genes within the parental chromosome length range one by one, calculating the fitness, and picking out the highest fitness children individuals.
Experimental data show that such an improvement can reduce the number of iterations and speed up the convergence of fitness.
\begin{algorithm}[htb]
        \caption{Crossover with fitness as evaluation.}
        \label{alg:crossover}
        \begin{algorithmic}[1]
        \Require Father’s gene, mother’s gene, fitness function;
        \Ensure  Child’s gene;
        \Function {crossover}{$father, mother, fitness$}
                \State $best\_fitness = float.MIN\_VALUE$;
                \State $best\_child = np.zeros(father.size)$;
                \For{$i = 0 \to father.size$}
                        \State $current\_child = np.zeros(father.size)$;
                        \State $current\_child = np.append(father[0:i], mother[i:])$;
                        \State $current\_fitness = fitness(current\_child)$;
                        \If{$current\_fitness > best\_fitness$}
                                \State $best\_fitness = current\_fitness$;
                                \State $best\_child = current\_child.copy()$;
                        \EndIf
                \EndFor
        \State \Return{$best\_child$}
        \EndFunction
        \end{algorithmic}
\end{algorithm}

\subsection{Improved Mutation Operation}
As shown in algorithm \ref{alg:mutate} is the pseudocode of the improved mutation algorithm.
The simple mutation of SGA sets a relatively large mutation rate, and mutates any one gene of the incoming children's chromosome when the generated random number is smaller than the mutation rate.
In this paper, we improve the algorithm by setting a small mutation rate and then selectively mutating each gene of the incoming children's chromosome.
That is, when the generated random number is smaller than the mutation rate, the gene is mutated, and when the traversed gene position is larger than half of the chromosome length, the mutation rate is set to twice the original one (the second half of the gene has relatively less influence on the result).
This ensures that the first half of the gene and the second half of the gene have an equal chance of mutation respectively, and can mutate at the same time.
When the gene length is $784$, the mutation rate of the whole chromosome is $1-(1-0.025)^{392}×(1-0.05)^{392}$, which greatly improves the species diversity and at the same time ensures the stability of the species (in the stable situation of the local optimum improves the global search ability), and experimental data show that it can improve the search capability.
\begin{algorithm}[htb]
        \caption{Mutate child with alter each gene if rand number less than mutate rate.}
        \label{alg:mutate}
        \begin{algorithmic}[1]
        \Require Child’s gene;
        \Ensure  Mutated child’s gene;
        \Function {mutate}{$child$}
                \State $mutate\_rate = 0.025$;
                \For{$i = 0 \to child.size$}
                        \If{$i > child.size//2$}
                                \State $mutate\_rate = 0.05$;
                        \EndIf
                        \If{$random.random() < mutate\_rate$}
                                \State $child[i] = !child[i]$;//child[i] equals 0 or 1
                        \EndIf
                \EndFor
        \State \Return{$child$}
        \EndFunction
        \end{algorithmic}
\end{algorithm}

\section{Numerical Experiments and Analysis}
\subsection{Test Functions}
In order to evaluate the optimization performance of the proposed improved genetic algorithm, 15 representative test functions from AVOA paper of \cite{abdollahzadeh2021african} and \cite{wikipedia2021test} are selected in this paper.
Since the proposed improved genetic algorithm is mainly used for the neural network adversarial attack problem, and the neural network has multi-dimensional parameters, the dimensions of the test functions will be tested on 30, 50, and 100, respectively.
The details of the formula, dimensions, range, and minimum of the 15 test functions are shown in \tablename~\ref{tab:fun1}, \tablename~\ref{tab:fun2}, and \tablename~\ref{tab:fun3}, where \tablename~\ref{tab:fun1} are multi-dimensional test functions with unimodal, \tablename~\ref{tab:fun2} are multi-dimensional test functions with multi-modal, and \tablename~\ref{tab:fun3} for fixed-dimensional test functions.

\begin{table}[htb]
        \caption{Details of unimodal test functions}
        \label{tab:fun1}
        \begin{tabular}{@{}lllll@{}}
        \toprule
        No & Function                                                                                           & Dimensions & Range            & $F_{min}$ \\ \midrule
        F1 & $f(x)=\sum_{i=1}^{d} x_{i}^{2}$                                                                    & 30,50,100  & $[-100,100]^{d}$ & 0         \\
        F2 & $f(x)=\sum_{i=1}^{d}\left|x_{i}\right|+\prod_{i=1}^{d}\left|x_{i}\right|$                          & 30,50,100  & $[-10,10]^{d}$   & 0         \\
        F3 & $f(x)=\sum_{i=1}^{d}\left(\sum_{j=1}^{i} x_{j}\right)^{2}$                                         & 30,50,100  & $[-100,100]^{d}$ & 0         \\
        F4 & $f(x)=\max _{i}\left\{\left|x_{i}\right|, 1 \leq i \leq d\right\}$                                 & 30,50,100  & $[-100,100]^{d}$ & 0         \\
        F5 & $f(x)=\sum_{i=1}^{d-1}\left[100\left(x_{i+1}-x_{i}^{2}\right)^{2}+\left(x_{i}-1\right)^{2}\right]$ & 30,50,100  & $[-30,30]^{d}$   & 0         \\
        F6 & $f(x)=\sum_{i=1}^{d}\left(\left|x_{i}+0.5\right|\right)^{2}$                                       & 30,50,100  & $[-100,100]^{d}$ & 0         \\
        F7 & $f(x)=\sum_{i=1}^{d} i x_{i}^{4}+\operatorname{random}[0,1)$                                       & 30,50,100  & $[-128,128]^{d}$ & 0         \\ \bottomrule
        \end{tabular}
\end{table}

\begin{table}[htb]
        \caption{Details of multi-modal test functions}
        \label{tab:fun2}
        \resizebox{\textwidth}{!}{
        \begin{tabular}{@{}lllll@{}}
        \toprule
        No  & Function                                                                                                                                         & Dimensions & Range              & $F_{min}$    \\ \midrule
        F8  & $f(x)=-\sum_{i=1}^{d}\left(x_{i} \sin \left(\sqrt{\left|x_{i}\right|}\right)\right)$                                                             & 30,50,100  & $[-500,500]^{d}$   & $−418.9829d$ \\
        F9  & $f(x)=10 d+\sum_{i=1}^{d}\left[x_{i}^{d}-10 \cos \left(2 \pi x_{i}\right)\right]$                                                                & 30,50,100  & $[-5.12,5.12]^{d}$ & 0            \\
        F10 & $f(x)=-20 \exp \left(-0.2 \sqrt{\frac{1}{d} \sum_{i=1}^{d} x_{i}^{2}}\right)-\exp \left(\frac{1}{d} \sum_{i=1}^{d} \cos 2 \pi x_{i}\right)+20+e$ & 30,50,100  & $[-32,32]^{d}$     & 0            \\
        F11 & $f(x)=\frac{1}{4000} \sum_{i=1}^{d} x_{i}^{2}-\prod_{i=1}^{d} \cos \left(\frac{x_{i}}{\sqrt{i}}\right)+1$                                        & 30,50,100  & $[-600,600]^{d}$   & 0            \\ \bottomrule
        \end{tabular}
        }
\end{table}

\begin{table}[htb]
        \caption{Details of fixed-dimension test functions}
        \label{tab:fun3}
        \resizebox{\textwidth}{!}{
        \begin{tabular}{@{}lllll@{}}
        \toprule
        No  & Function                                                                                                           & Dimensions & Range              & $F_{min}$ \\ \midrule
        F12 & $f(x, y)=x^{2}+y^{2}$                                                                                              & 2          & $[-5,5]^{x,y}$     & 0         \\
        F13 & $f(x, y)=-20 \exp \left[-0.2 \sqrt{0.5\left(x^{2}+y^{2}\right)}\right]-\exp [0.5(\cos 2 \pi x+\cos 2 \pi y)]+e+20$ & 2          & $[-5,5]^{x,y}$     & 0         \\
        F14 & $f(x, y)=(1.5-x+x y)^{2}+\left(2.25-x+x y^{2}\right)^{2}+\left(2.625-x+x y^{3}\right)^{2}$                         & 2          & $[-4.5,4.5]^{x,y}$ & 0         \\
        F15 & $f(x, y)=-(y+47) \sin \sqrt{\left|\frac{x}{2}+(y+47)\right|}-x \sin \sqrt{|x-(y+47)|}$                             & 2          & $[-512,512]^{x,y}$ & −959.6407 \\ \bottomrule
        \end{tabular}
        }
\end{table}

\subsection{Experimental Environment}
The hardware environment of the experiment includes 8G of RAM, i7-4700MQ CPU; the software environment includes Windows 10 system, and the version of Python is 3.8.8.
In order to compare the optimization performance of IGA, SGA (Simple Genetic Algorithm), PSO (Particle Swarm Optimization) and GWO (Grey Wolf Optimizer) are selected as the experimental objects for comparison experiments in this paper.

\begin{figure}[ht]
        \centering
        \subfigure[Mutation rate]{
                \includegraphics[width=0.3\textwidth]{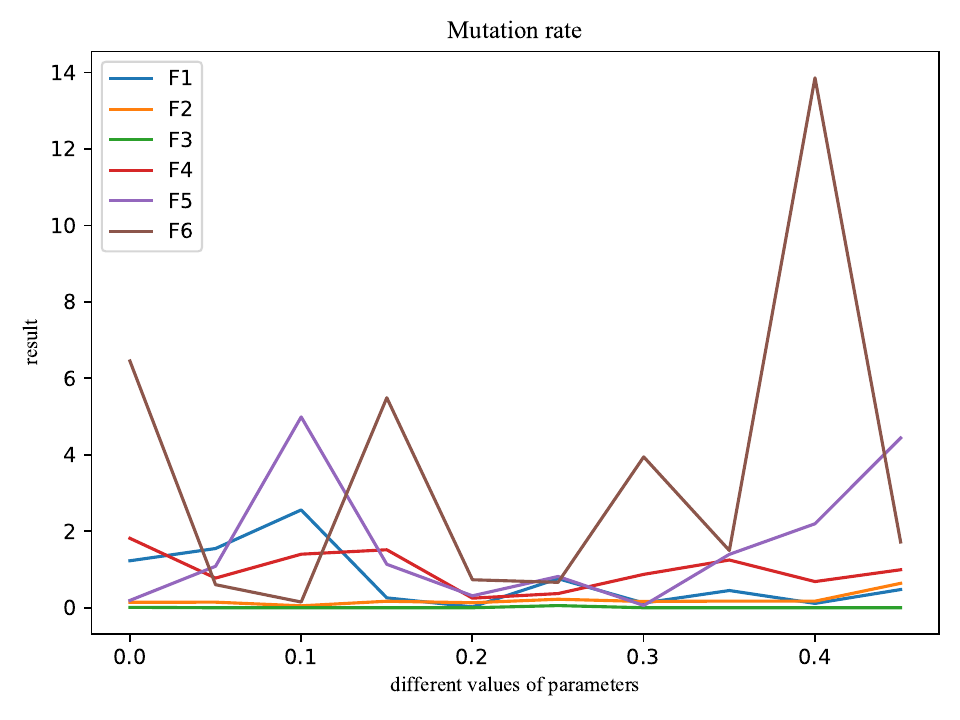}
        }
        \subfigure[Population size]{
                \includegraphics[width=0.3\textwidth]{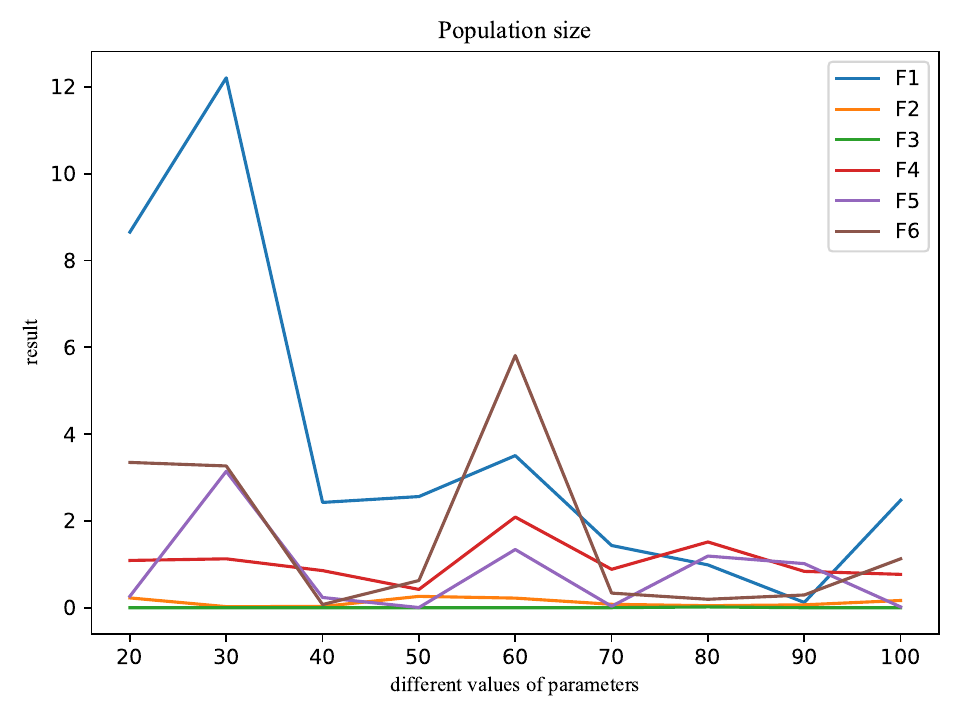}
        }
        \subfigure[Max iteration]{
                \includegraphics[width=0.3\textwidth]{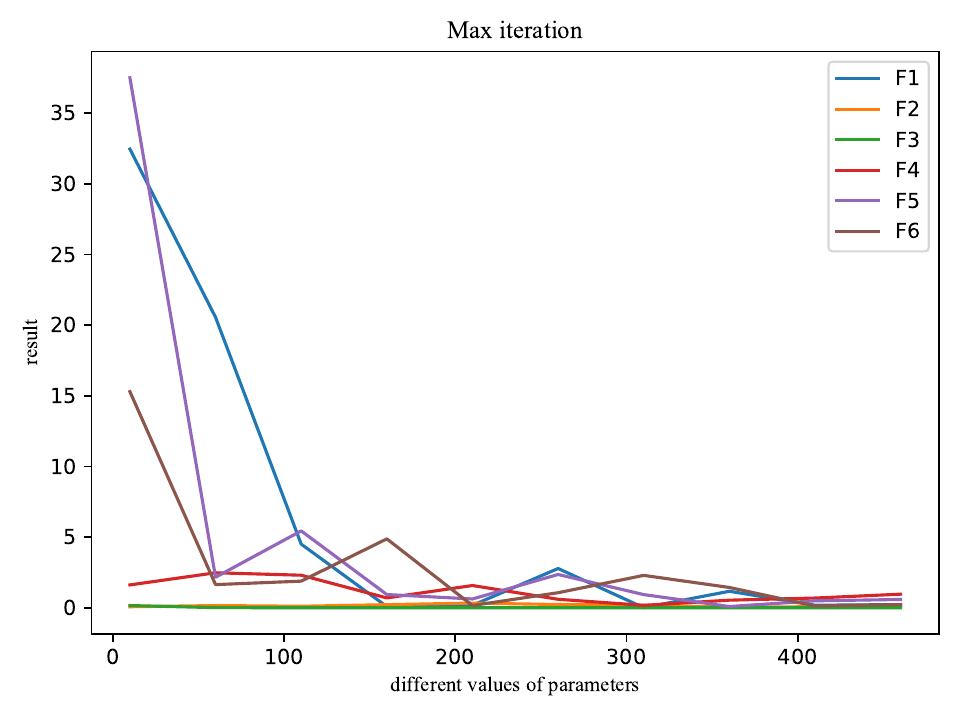}
        }
        \caption{IGA parameters selection}
        \label{fig:param_sele}
\end{figure}

As shown in \figurename~\ref{fig:param_sele}, in order to determine the appropriate parameters for the IGA, this paper combines different parameters of the IGA and then tests them several times on F1-F6.
The detailed parameters of the 4 optimization algorithms are shown in \tablename~\ref{tab2}, and the population size and the max iteration are kept the same for the convenience of comparison.
The other parameters of PSO are set to typical values: $w = 1$, $c_1 = c_2 = 1.49445$.
The other parameters of GWO are set to typical values: $\vec{C}=\operatorname{Rand}(0,2)$, $\vec{a}=\operatorname{Rand}(-a,a)$, $a=2 \to 0$.

\begin{table}[!htp]
        \centering
        \caption{The parameter settings}
        \label{tab2}
        \begin{tabular}{ccccc}
                \toprule
                Algorithm & Iteration & Population size & Gene length & Mutation rate \\
                \midrule
                IGA & 101 & 50 & 30 & 0.05 \\
                SGA & 101 & 50 & 30 & 0.2 \\
                PSO & 101 & 50 & - & - \\
                GWO & 101 & 50 & - & - \\
                \bottomrule
        \end{tabular}
\end{table}

\subsection{Experimental Results and Analysis}
\subsubsection{Qualitative Result Analysis}
\begin{figure}[ht]
        \centering
        \subfigure[Parameter space]{
                \includegraphics[width=0.22\textwidth]{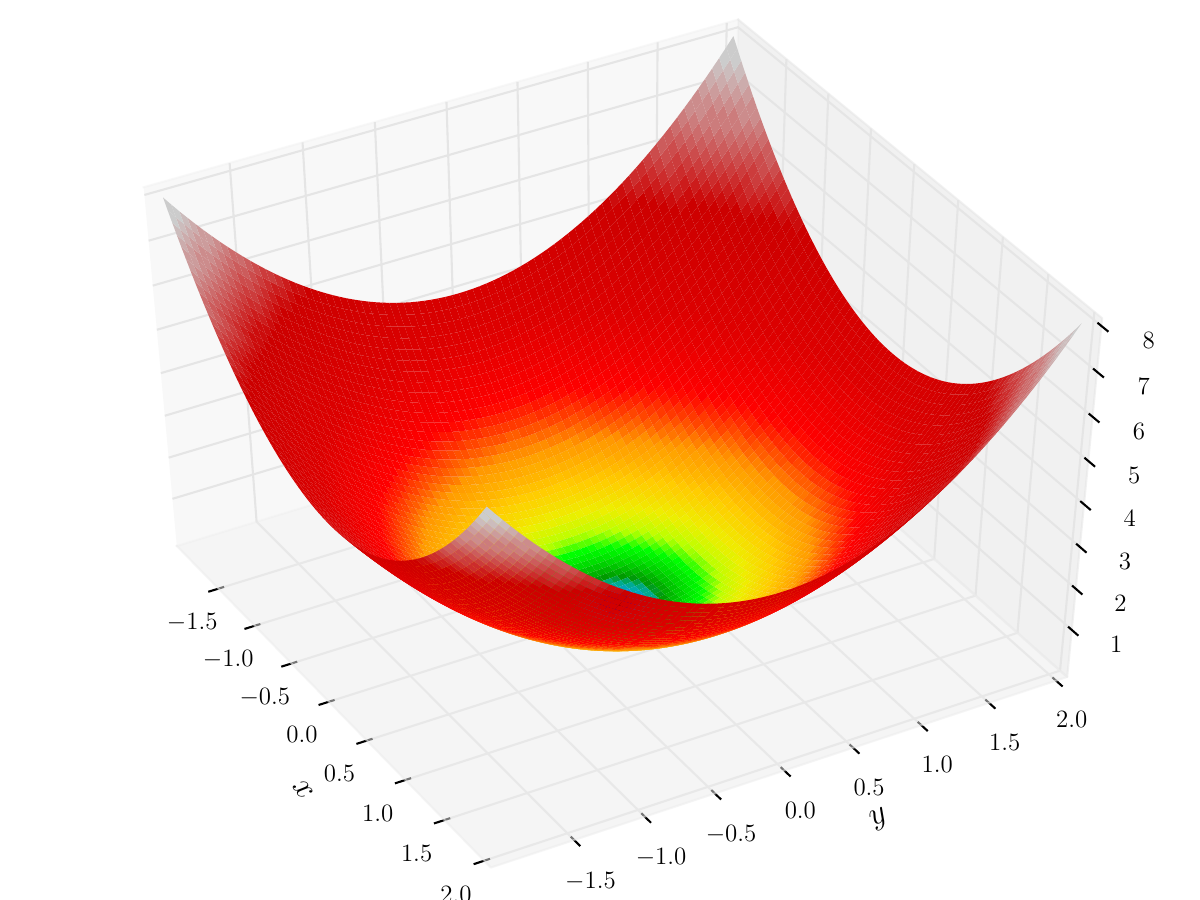}
        }
        \subfigure[Population distribution]{
                \includegraphics[width=0.22\textwidth]{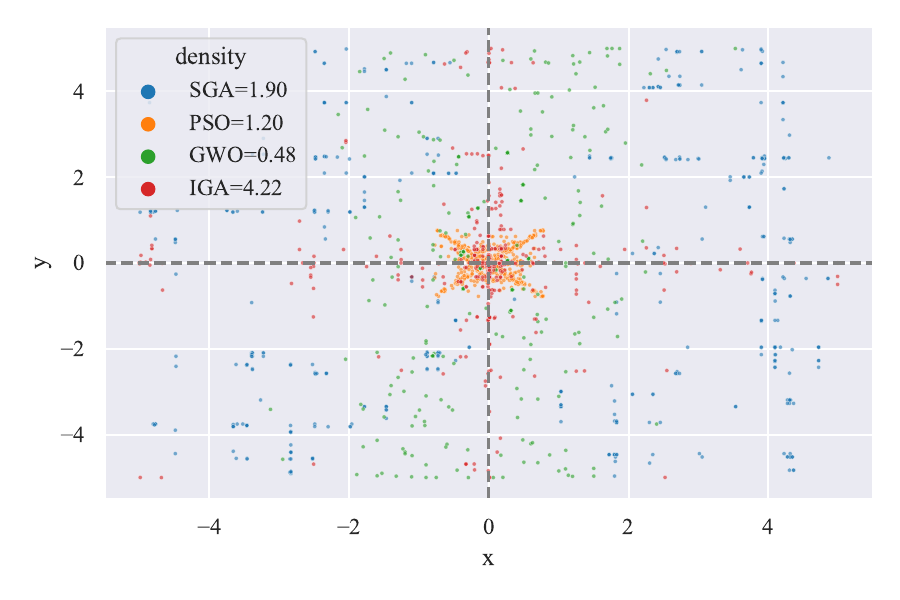}
        }
        \subfigure[Best record]{
                \includegraphics[width=0.22\textwidth]{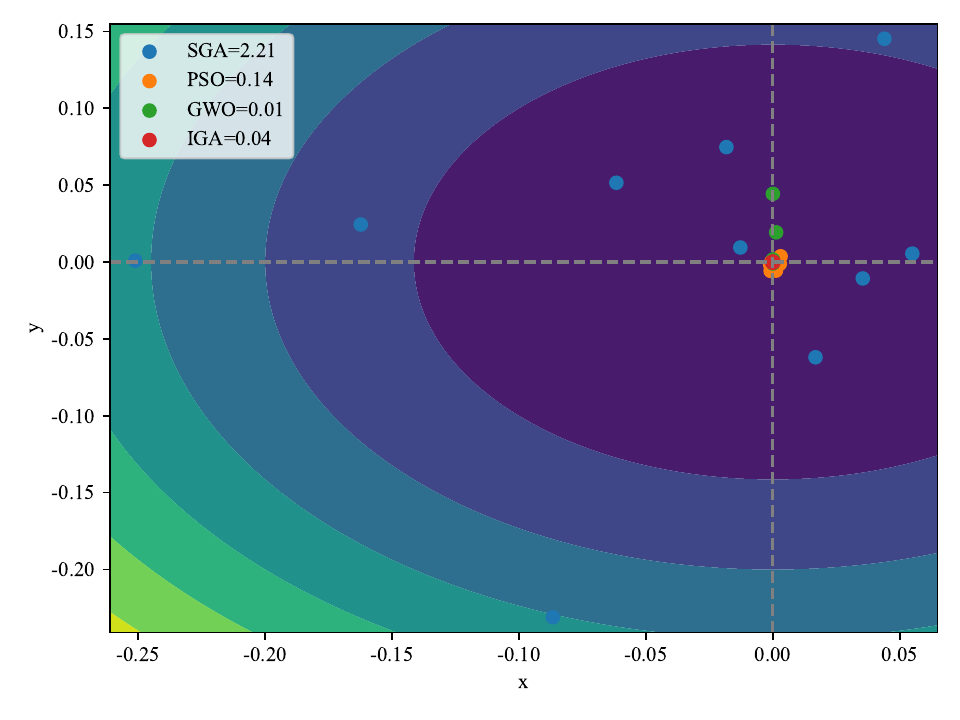}
        }
        \subfigure[Convergence curve]{
                \includegraphics[width=0.22\textwidth]{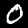}
        }
        \caption{Qualitative results for the F12 function}
        \label{fig:f12}
\end{figure}

\begin{figure}[ht]
        \centering
        \subfigure[Parameter space]{
                \includegraphics[width=0.22\textwidth]{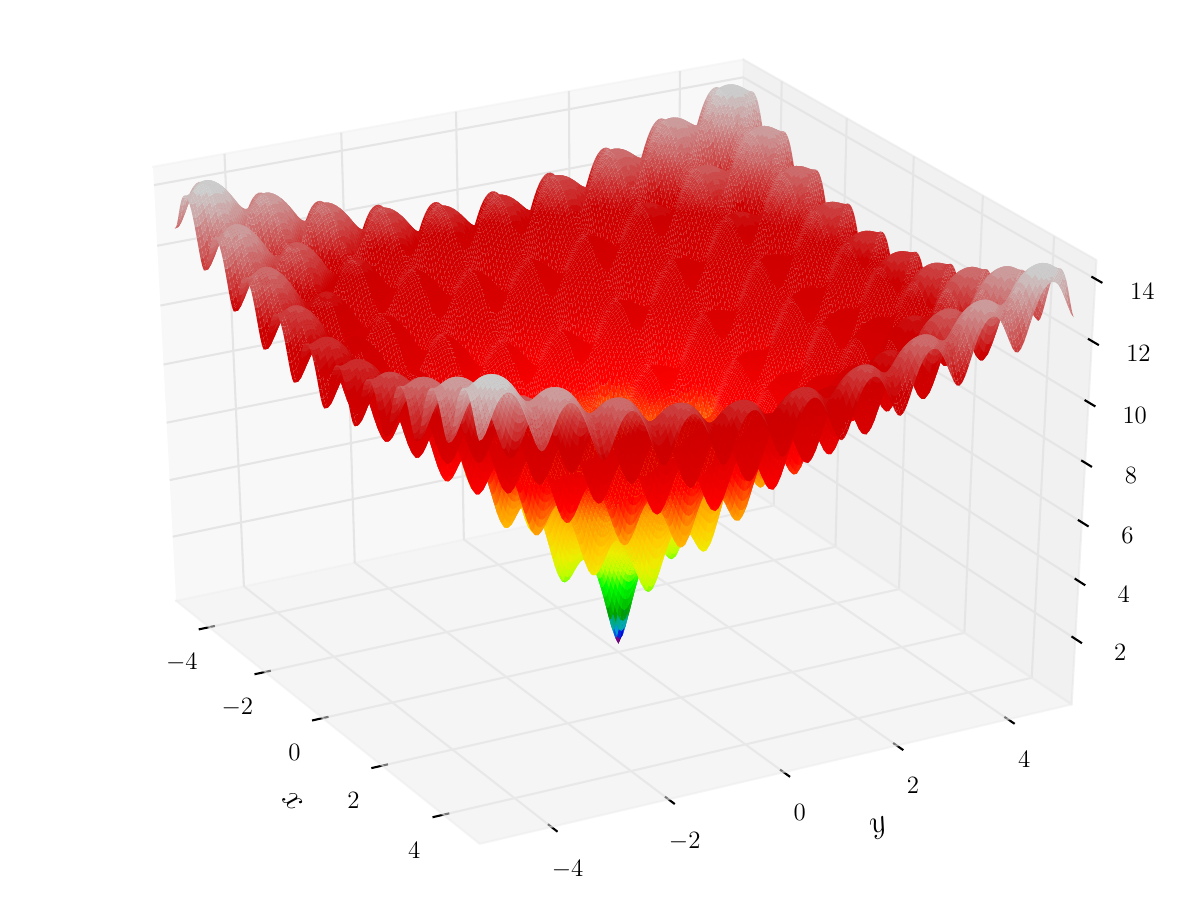}
        }
        \subfigure[Population distribution]{
                \includegraphics[width=0.22\textwidth]{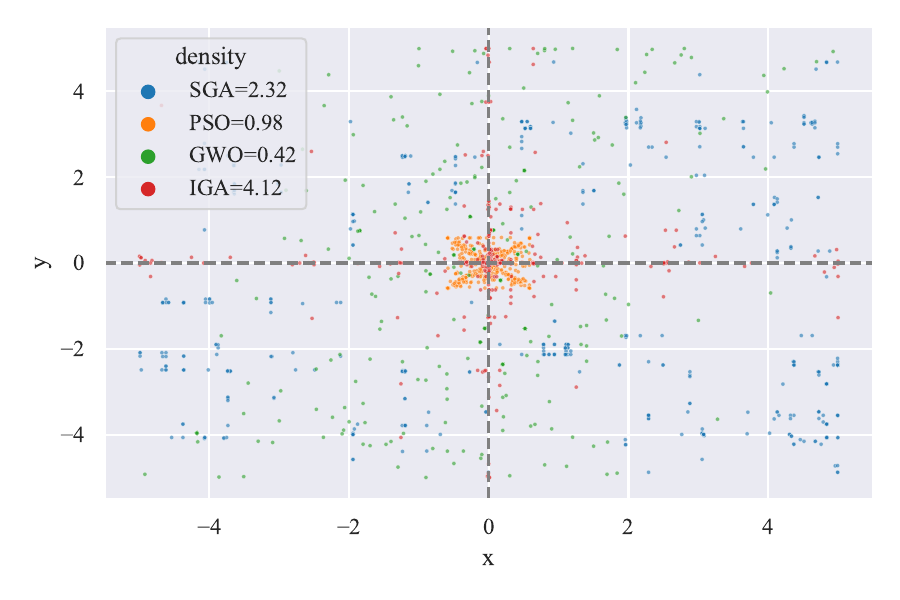}
        }
        \subfigure[Best record]{
                \includegraphics[width=0.22\textwidth]{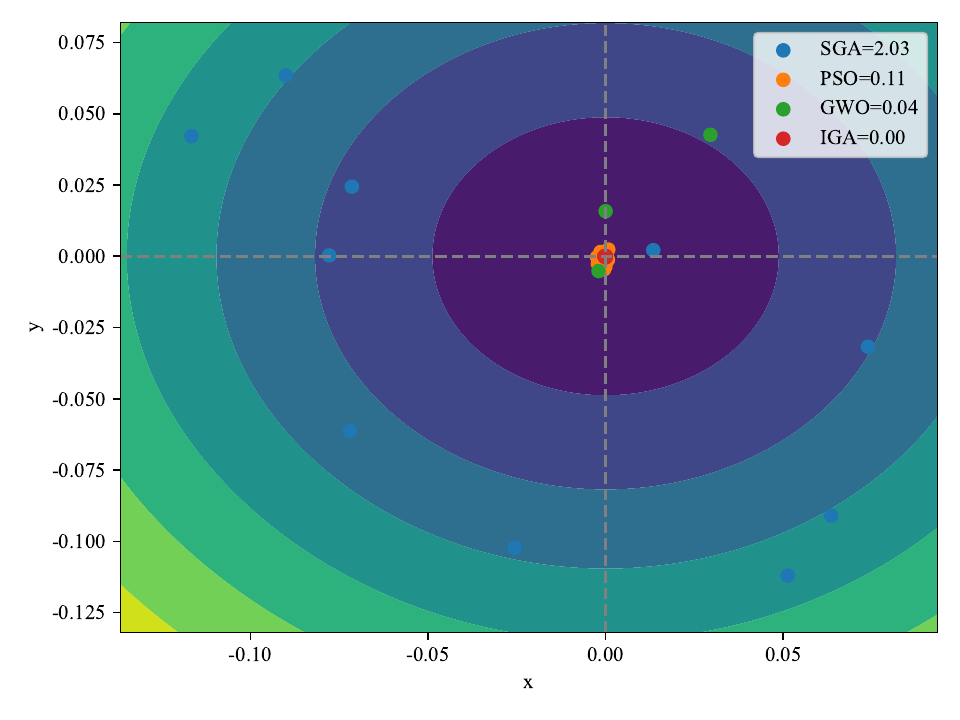}
        }
        \subfigure[Convergence curve]{
                \includegraphics[width=0.22\textwidth]{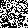}
        }
        \caption{Qualitative results for the F13 function}
        \label{fig:f13}
\end{figure}

\begin{figure}[ht]
        \centering
        \subfigure[Parameter space]{
                \includegraphics[width=0.22\textwidth]{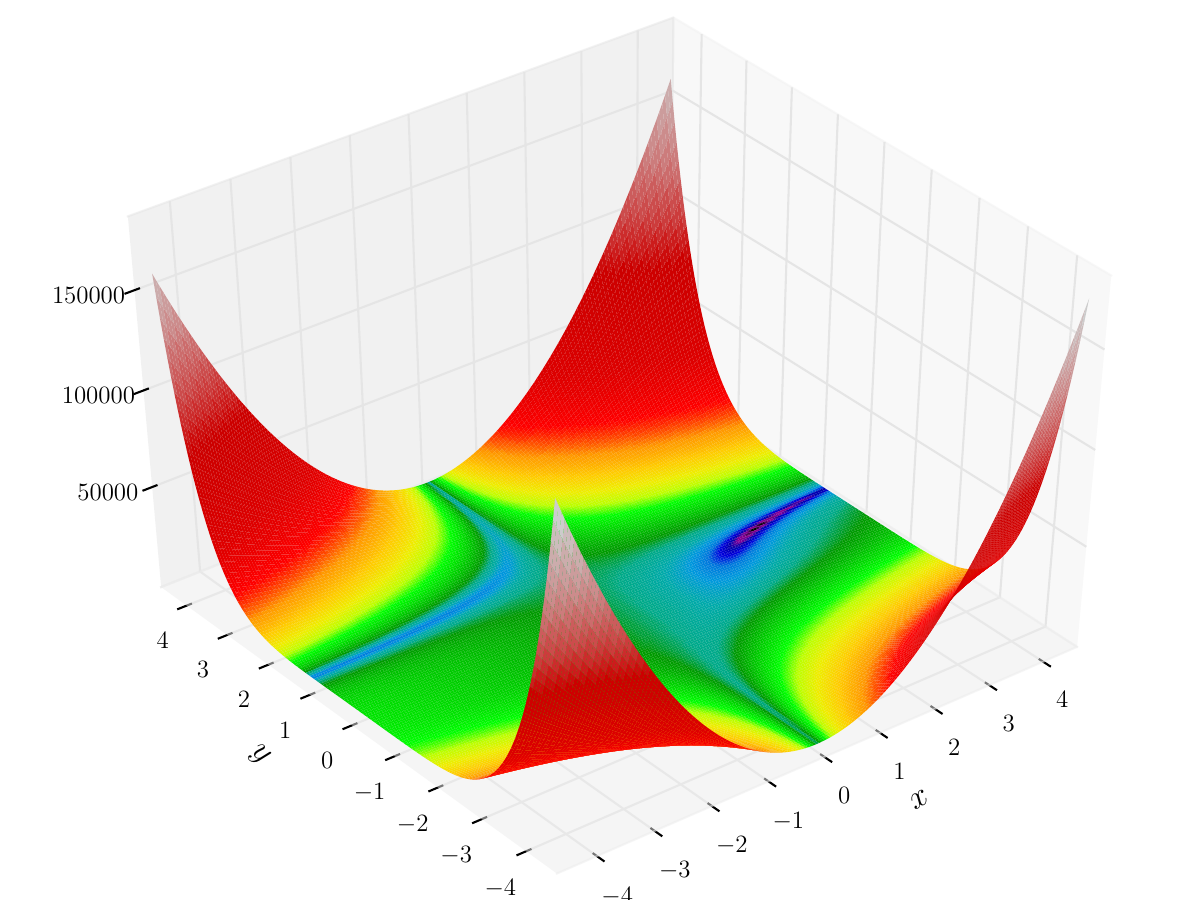}
        }
        \subfigure[Population distribution]{
                \includegraphics[width=0.22\textwidth]{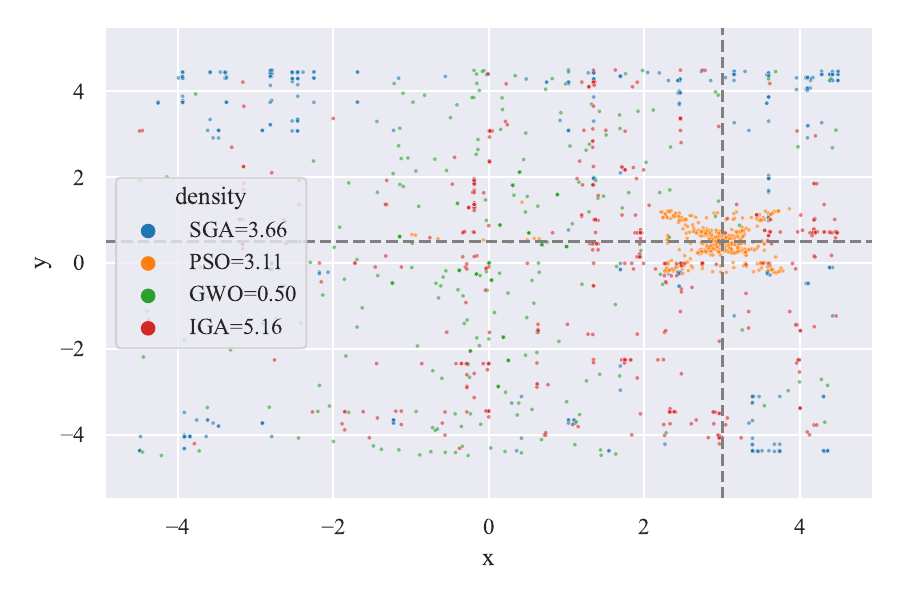}
        }
        \subfigure[Best record]{
                \includegraphics[width=0.22\textwidth]{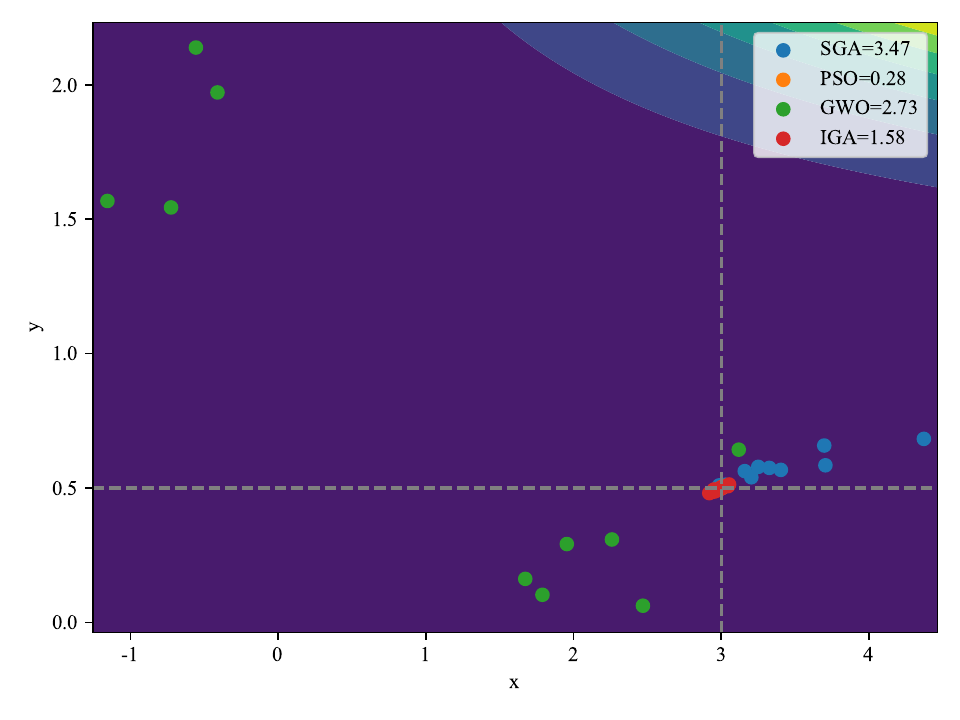}
        }
        \subfigure[Convergence curve]{
                \includegraphics[width=0.22\textwidth]{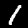}
        }
        \caption{Qualitative results for the F14 function}
        \label{fig:f14}
\end{figure}

\begin{figure}[ht]
        \centering
        \subfigure[Parameter space]{
                \includegraphics[width=0.22\textwidth]{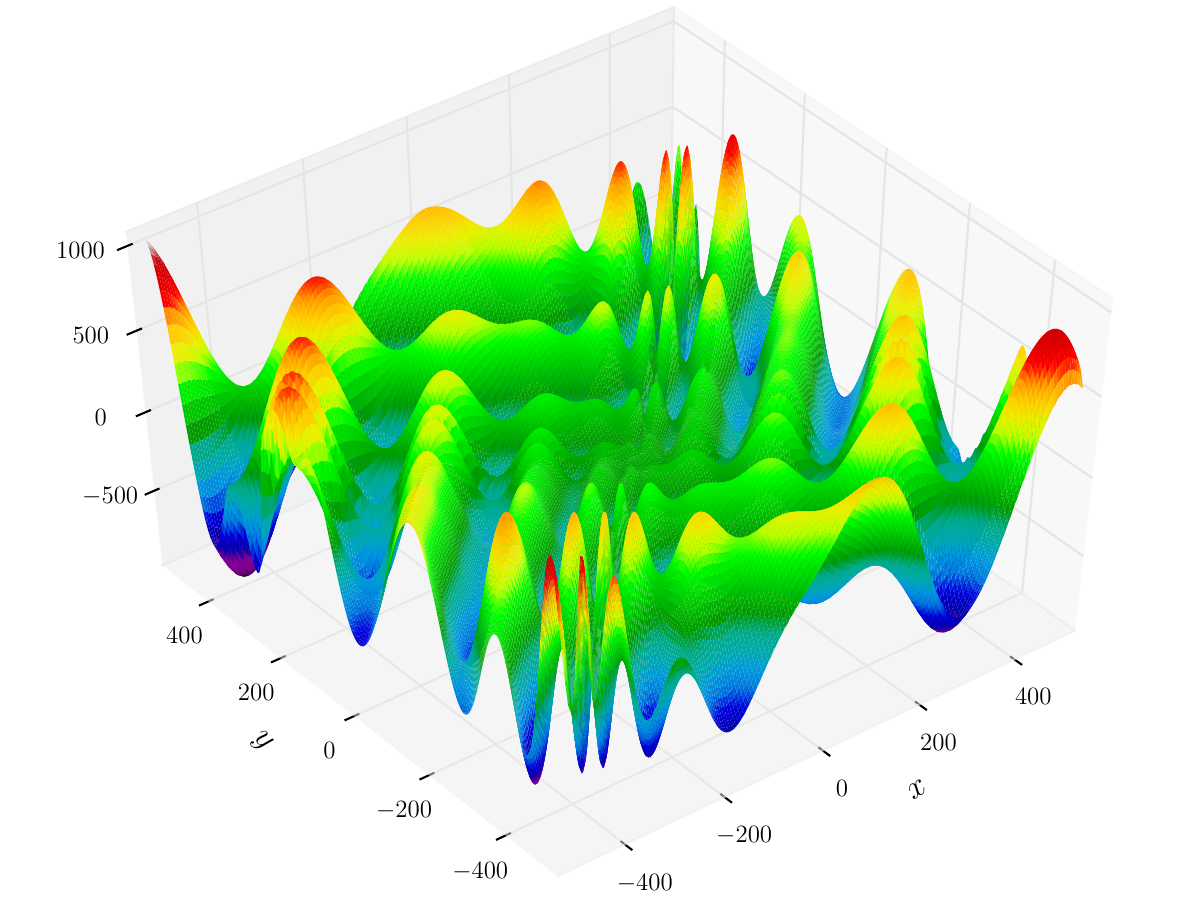}
        }
        \subfigure[Population distribution]{
                \includegraphics[width=0.22\textwidth]{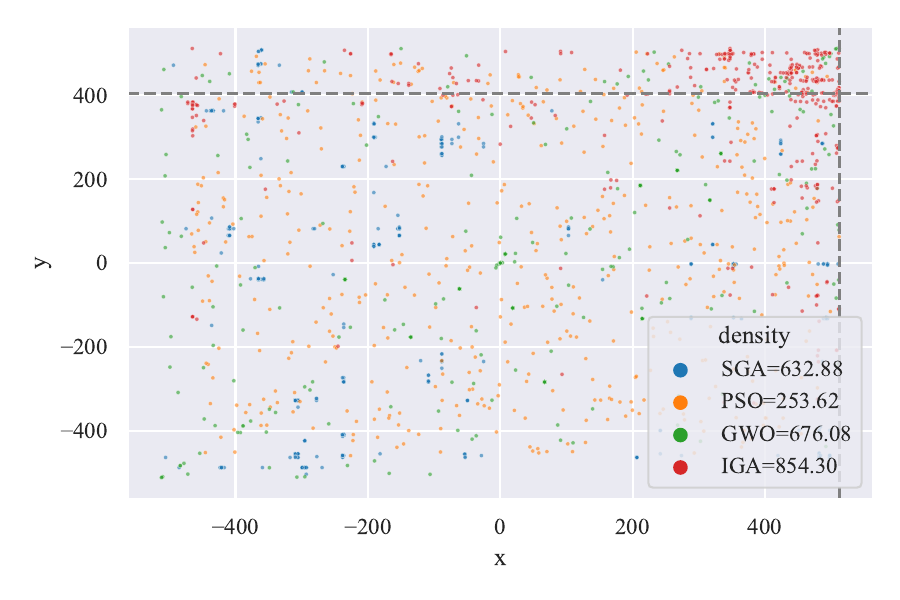}
        }
        \subfigure[Best record]{
                \includegraphics[width=0.22\textwidth]{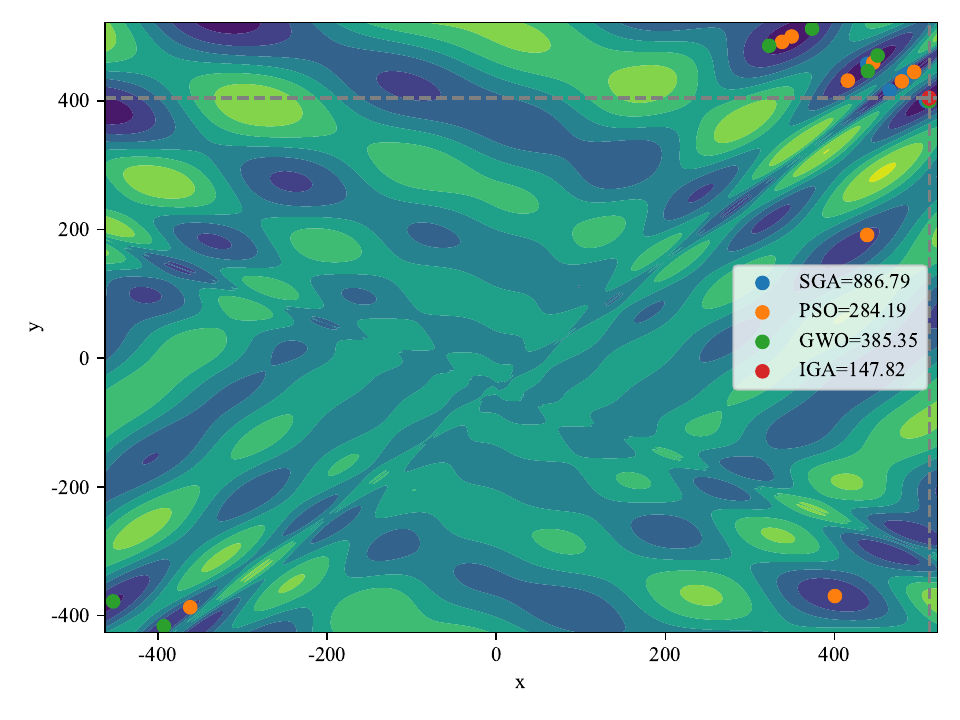}
        }
        \subfigure[Convergence curve]{
                \includegraphics[width=0.22\textwidth]{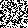}
        }
        \caption{Qualitative results for the F15 function}
        \label{fig:f15}
\end{figure}

As shown in \figurename~\ref{fig:f12}, \figurename~\ref{fig:f13}, \figurename~\ref{fig:f14} and \figurename~\ref{fig:f15}, F12-F15 are used to evaluate the qualitative results of the IGA.
Each optimization algorithm was tested 10 times on F12-F15 under the same experimental conditions.
Among them, "Population distribution" is the scatter plot of the distribution of all individuals for each optimization algorithm in 10 experiments, and the formula for the density is shown in Formula \eqref{eq6}, $population\_size = 50$.
"Best record" is the scatter plot of the distribution of the optimal individuals for each experiment, and the formula for calculating the intensity is shown in Formula \eqref{eq7}.
From the figure, we can see that the density of optimal individuals for each round of experimental IGA is better than the other three optimization algorithms, and also retains a strong global search capability in the last iteration.
On the F15 test function, SGA, PSO and GWO fall into local optimum several times.
From the convergence curves, we can see that IGA is converged before the other three optimization algorithms, and the precision after convergence is better.

\begin{equation}
        \text { density }=\frac{1}{population\_size} \sum_{i=1}^{population\_size} \operatorname{dist}\left(a_{i}, o\right)
        \label{eq6}
\end{equation}

\begin{equation}
        \text { density }=\frac{1}{test\_n} \sum_{i=1}^{test\_n} \operatorname{dist}\left(a_{best}, o\right)
        \label{eq7}
\end{equation}

\subsubsection{Quantitative Result Analysis}
In order to make a quantitative comparison with the other three mainstream optimization algorithms, the four optimization algorithms are performed independently for 10 experiments on F1-F11 test functions in dimensions 30, 50, and 100, respectively.
The purpose of performing the high-dimensional function test is to test the convergence superiority of IGA on the high-dimensional space for application in the field of neural network adversarial attack.
\tablename~\ref{tab:res1}, \tablename~\ref{tab:res2}, \tablename~\ref{tab:res3} are the test results of the test functions F1-F11 in 30, 50, and 100 dimensions, respectively.
\tablename~\ref{tab:res4} shows the results of the four optimization algorithms tested on the test functions F12-F15.
The best result, worst result, mean, median, standard deviation, and P-value are compared for 10 experiments.
Where P-value is the result of the Wilcoxon rank-sum statistical test and P-value below $5\%$ is significant.

In \tablename~\ref{tab:res1}, IGA achieves significantly superior performance in 9 test functions, PSO is better in F3, and SGA is slightly better in F8.
In \tablename~\ref{tab:res2} and \tablename~\ref{tab:res3}, IGA achieves significantly superior performance in 10 test functions, PSO performs better in F3.
It can be seen that the performance loss of IGA with increasing dimensionality is not as large as the other three optimization algorithms.
In \tablename~\ref{tab:res4}, IGA achieves significantly superior performance in 3 test functions, and PSO performs slightly better in F14.

In general, IGA has better iteration efficiency, global search capability, and convergence success rate than the other three optimization algorithms.

\begin{table}
\centering
\caption{Results of test functions (F1-F11) with 30 dimensions}
\label{tab:res1}
\resizebox{\textwidth}{!}{
        \begin{tabular}{@{}llllllll@{}}
        \toprule
        Fun & Alg & Min                & Max                & Mean               & Median             & Std               & P-value           \\ \midrule
        F1  & SGA & 5.60E+03           & 1.24E+04           & 8.29E+03           & 7.68E+03           & 2.44E+03          & 1.57E-04          \\
            & PSO & 2.14E+02           & 2.87E+02           & 2.48E+02           & 2.45E+02           & 2.65E+01          & 1.57E-04          \\
            & GWO & 2.64E-15           & 4.54E-14           & 1.81E-14           & 1.58E-14           & 1.56E-14          & 1.57E-04          \\
            & IGA & \textbf{1.75E-18}  & \textbf{5.12E-16}  & \textbf{7.44E-17}  & \textbf{1.10E-17}  & \textbf{1.60E-16} & 1.00E+00          \\
        F2  & SGA & 8.15E+01           & 1.44E+05           & 2.08E+04           & 9.32E+02           & 4.46E+04          & 1.57E-04          \\
            & PSO & 8.11E+01           & 8.77E+05           & 1.49E+05           & 1.45E+04           & 2.94E+05          & 1.57E-04          \\
            & GWO & 6.63E-12           & 7.78E-10           & 2.41E-10           & 8.51E-11           & 2.95E-10          & 3.81E-04          \\
            & IGA & \textbf{2.18E-12}  & \textbf{1.05E-11}  & \textbf{5.20E-12}  & \textbf{4.92E-12}  & \textbf{2.91E-12} & 1.00E+00          \\
        F3  & SGA & 2.72E+04           & 5.50E+04           & 4.09E+04           & 4.11E+04           & 8.25E+03          & 1.57E-04          \\
            & PSO & \textbf{1.92E+02}  & \textbf{3.12E+02}  & \textbf{2.33E+02}  & \textbf{2.26E+02}  & \textbf{3.46E+01} & 1.57E-04          \\
            & GWO & 7.20E+03           & 1.75E+04           & 1.14E+04           & 1.07E+04           & 3.89E+03          & 4.13E-02          \\
            & IGA & 3.07E+03           & 2.10E+04           & 8.39E+03           & 6.87E+03           & 5.11E+03          & 1.00E+00          \\
        F4  & SGA & 5.92E+01           & 7.54E+01           & 6.74E+01           & 6.65E+01           & 5.52E+00          & 1.57E-04          \\
            & PSO & 5.00E+00           & 5.00E+00           & 5.00E+00           & 5.00E+00           & 0.00E+00          & 1.57E-04          \\
            & GWO & 1.62E-02           & 1.13E+00           & 3.15E-01           & 1.58E-01           & 4.04E-01          & 1.57E-04          \\
            & IGA & \textbf{1.18E-05}  & \textbf{9.82E-05}  & \textbf{3.94E-05}  & \textbf{2.64E-05}  & \textbf{2.68E-05} & 1.00E+00          \\
        F5  & SGA & 9.92E+06           & 2.47E+07           & 1.65E+07           & 1.46E+07           & 5.43E+06          & 1.57E-04          \\
            & PSO & 3.02E+05           & 4.82E+05           & 3.94E+05           & 3.95E+05           & 5.53E+04          & 1.57E-04          \\
            & GWO & 2.88E+01           & \textbf{2.88E+01}  & 2.88E+01           & 2.88E+01           & \textbf{1.10E-02} & 1.57E-04          \\
            & IGA & \textbf{2.80E+01}  & \textbf{2.88E+01}  & \textbf{2.87E+01}  & \textbf{2.87E+01}  & 2.23E-01          & 1.00E+00          \\
        F6  & SGA & 4.22E+03           & 1.38E+04           & 9.91E+03           & 9.68E+03           & 3.22E+03          & 1.57E-04          \\
            & PSO & 1.88E+02           & 2.76E+02           & 2.41E+02           & 2.40E+02           & 2.70E+01          & 1.57E-04          \\
            & GWO & 9.70E-01           & 3.24E+00           & 2.26E+00           & 2.47E+00           & 7.05E-01          & 1.50E-03          \\
            & IGA & \textbf{6.69E-01}  & \textbf{1.53E+00}  & \textbf{1.05E+00}  & \textbf{1.02E+00}  & \textbf{2.93E-01} & 1.00E+00          \\
        F7  & SGA & 3.60E+08           & 1.09E+09           & 7.13E+08           & 6.67E+08           & 2.27E+08          & 1.57E-04          \\
            & PSO & 3.51E+04           & 5.59E+04           & 4.70E+04           & 4.68E+04           & 7.23E+03          & 1.57E-04          \\
            & GWO & 8.06E-04           & 4.88E-02           & 1.51E-02           & 7.92E-03           & 1.71E-02          & 6.50E-03          \\
            & IGA & \textbf{3.45E-04}  & \textbf{8.24E-03}  & \textbf{2.20E-03}  & \textbf{1.53E-03}  & \textbf{2.39E-03} & 1.00E+00          \\
        F8  & SGA & \textbf{-9.28E+03} & -7.16E+03          & -8.68E+03          & -8.94E+03          & 6.92E+02          & \textbf{7.05E-01} \\
            & PSO & -1.18E+02          & -8.65E+01          & -1.04E+02          & -1.02E+02          & 8.93E+00          & 1.57E-04          \\
            & GWO & -9.11E+03          & -6.76E+03          & -8.50E+03          & -8.70E+03          & 7.20E+02          & \textbf{1.51E-01} \\
            & IGA & -9.02E+03          & \textbf{-8.60E+03} & \textbf{-8.90E+03} & \textbf{-9.01E+03} & \textbf{1.67E+02} & 1.00E+00          \\
        F9  & SGA & 3.65E+13           & 5.31E+17           & 7.97E+16           & 5.50E+15           & 1.66E+17          & 1.57E-04          \\
            & PSO & 1.33E+21           & 3.50E+21           & 2.47E+21           & 2.42E+21           & 7.43E+20          & 1.57E-04          \\
            & GWO & 1.14E-13           & 2.22E+02           & 3.48E+01           & 8.81E-13           & 7.67E+01          & 1.83E-04          \\
            & IGA & \textbf{0.00E+00}  & \textbf{1.14E-13}  & \textbf{2.27E-14}  & \textbf{0.00E+00}  & \textbf{3.97E-14} & 1.00E+00          \\
        F10 & SGA & 1.32E+01           & 1.63E+01           & 1.49E+01           & 1.52E+01           & 1.11E+00          & 1.57E-04          \\
            & PSO & 9.51E+00           & 1.08E+01           & 1.02E+01           & 1.03E+01           & 4.61E-01          & 1.57E-04          \\
            & GWO & 2.62E-08           & 5.37E-07           & 1.88E-07           & 1.21E-07           & 1.78E-07          & 1.57E-04          \\
            & IGA & \textbf{5.71E-11}  & \textbf{4.23E-09}  & \textbf{9.95E-10}  & \textbf{5.75E-10}  & \textbf{1.27E-09} & 1.00E+00          \\
        F11 & SGA & 5.66E+01           & 1.35E+02           & 9.91E+01           & 1.01E+02           & 2.20E+01          & 1.57E-04          \\
            & PSO & 1.05E+00           & 1.07E+00           & 1.06E+00           & 1.06E+00           & 4.80E-03          & 1.57E-04          \\
            & GWO & 4.22E-15           & 2.88E-08           & 2.88E-09           & 1.08E-12           & 9.10E-09          & 1.57E-04          \\
            & IGA & \textbf{0.00E+00}  & \textbf{1.33E-15}  & \textbf{2.22E-16}  & \textbf{1.11E-16}  & \textbf{4.05E-16} & 1.00E+00          \\ \bottomrule
        \end{tabular}
}
\end{table}

\begin{table}
\centering
\caption{Results of test functions (F1-F11) with 50 dimensions}
\label{tab:res2}
\resizebox{\textwidth}{!}{
        \begin{tabular}{@{}llllllll@{}}
        \toprule
        Fun & Alg & Min                & Max                & Mean               & Median             & Std               & P-value  \\ \midrule
        F1  & SGA & 1.54E+04           & 3.39E+04           & 2.69E+04           & 2.86E+04           & 5.70E+03          & 1.57E-04 \\
            & PSO & 4.92E+02           & 5.53E+02           & 5.22E+02           & 5.25E+02           & 2.26E+01          & 1.57E-04 \\
            & GWO & 6.08E-16           & 8.65E-11           & 9.74E-12           & 1.14E-12           & 2.70E-11          & 1.57E-04 \\
            & IGA & \textbf{3.91E-18}  & \textbf{3.10E-16}  & \textbf{8.88E-17}  & \textbf{6.02E-17}  & \textbf{1.03E-16} & 1.00E+00 \\
        F2  & SGA & 3.12E+09           & 8.84E+15           & 1.78E+15           & 7.55E+12           & 3.45E+15          & 1.57E-04 \\
            & PSO & 4.25E+10           & 1.11E+14           & 1.64E+13           & 4.80E+12           & 3.37E+13          & 1.57E-04 \\
            & GWO & 1.54E-11           & 1.71E-09           & 8.26E-10           & 8.72E-10           & 5.82E-10          & 8.81E-04 \\
            & IGA & \textbf{2.57E-12}  & \textbf{1.03E-10}  & \textbf{3.02E-11}  & \textbf{2.04E-11}  & \textbf{3.19E-11} & 1.00E+00 \\
        F3  & SGA & 7.50E+04           & 1.50E+05           & 1.14E+05           & 1.17E+05           & 2.34E+04          & 1.57E-04 \\
            & PSO & \textbf{4.45E+02}  & \textbf{7.54E+02}  & \textbf{5.70E+02}  & \textbf{5.49E+02}  & \textbf{1.00E+02} & 1.57E-04 \\
            & GWO & 1.90E+04           & 6.22E+04           & 4.07E+04           & 4.01E+04           & 1.26E+04          & \textbf{1.74E-01} \\
            & IGA & 1.34E+04           & 4.91E+04           & 3.15E+04           & 3.20E+04           & 1.16E+04          & 1.00E+00 \\
        F4  & SGA & 6.64E+01           & 8.39E+01           & 7.60E+01           & 7.58E+01           & 5.34E+00          & 1.57E-04 \\
            & PSO & 5.00E+00           & 5.00E+00           & 5.00E+00           & 5.00E+00           & 0.00E+00          & 1.57E-04 \\
            & GWO & 9.01E-03           & 3.25E+01           & 4.85E+00           & 1.66E+00           & 9.96E+00          & 1.57E-04 \\
            & IGA & \textbf{3.86E-06}  & \textbf{1.92E-04}  & \textbf{7.04E-05}  & \textbf{6.60E-05}  & \textbf{5.94E-05} & 1.00E+00 \\
        F5  & SGA & 3.98E+07           & 8.59E+07           & 5.67E+07           & 5.65E+07           & 1.46E+07          & 1.57E-04 \\
            & PSO & 8.76E+05           & 1.04E+06           & 9.61E+05           & 9.59E+05           & 5.04E+04          & 1.57E-04 \\
            & GWO & 4.86E+01           & 4.87E+01           & 4.87E+01           & 4.87E+01           & 3.87E-02          & 8.81E-04 \\
            & IGA & \textbf{4.85E+01}  & \textbf{4.86E+01}  & \textbf{4.86E+01}  & \textbf{4.86E+01}  & \textbf{3.69E-02} & 1.00E+00 \\
        F6  & SGA & 1.90E+04           & 3.15E+04           & 2.46E+04           & 2.43E+04           & 3.26E+03          & 1.57E-04 \\
            & PSO & 4.74E+02           & 5.74E+02           & 5.24E+02           & 5.15E+02           & 3.62E+01          & 1.57E-04 \\
            & GWO & 2.58E+00           & 4.83E+00           & 3.77E+00           & 3.74E+00           & 6.99E-01          & 6.70E-04 \\
            & IGA & \textbf{1.87E+00}  & \textbf{3.55E+00}  & \textbf{2.43E+00}  & \textbf{2.36E+00}  & \textbf{5.01E-01} & 1.00E+00 \\
        F7  & SGA & 2.84E+09           & 6.44E+09           & 4.57E+09           & 4.53E+09           & 1.13E+09          & 1.57E-04 \\
            & PSO & 1.93E+05           & 2.51E+05           & 2.22E+05           & 2.21E+05           & 2.05E+04          & 1.57E-04 \\
            & GWO & 7.01E-03           & 5.29E-02           & 1.67E-02           & 1.20E-02           & 1.37E-02          & 1.57E-04 \\
            & IGA & \textbf{3.24E-04}  & \textbf{6.80E-03}  & \textbf{3.27E-03}  & \textbf{3.68E-03}  & \textbf{2.29E-03} & 1.00E+00 \\
        F8  & SGA & -1.35E+04          & -1.21E+04          & -1.31E+04          & -1.32E+04          & 4.14E+02          & 1.57E-04 \\
            & PSO & -1.65E+02          & -1.26E+02          & -1.43E+02          & -1.38E+02          & \textbf{1.52E+01} & 1.57E-04 \\
            & GWO & \textbf{-1.50E+04} & -1.03E+04          & -1.40E+04          & -1.44E+04          & 1.32E+03          & 1.15E-03 \\
            & IGA & \textbf{-1.50E+04} & \textbf{-1.47E+04} & \textbf{-1.49E+04} & \textbf{-1.50E+04} & 1.16E+02          & 1.00E+00 \\
        F9  & SGA & 5.17E+28           & 9.28E+30           & 2.40E+30           & 1.74E+30           & 2.69E+30          & 1.57E-04 \\
            & PSO & 5.69E+35           & 1.08E+36           & 8.10E+35           & 7.84E+35           & 1.70E+35          & 1.57E-04 \\
            & GWO & 2.27E-13           & 8.93E-10           & 1.20E-10           & 4.76E-11           & 2.73E-10          & 1.57E-04 \\
            & IGA & \textbf{0.00E+00}  & \textbf{1.14E-13}  & \textbf{2.27E-14}  & \textbf{0.00E+00}  & \textbf{4.79E-14} & 1.00E+00 \\
        F10 & SGA & 1.58E+01           & 1.82E+01           & 1.72E+01           & 1.72E+01           & 7.07E-01          & 1.57E-04 \\
            & PSO & 1.05E+01           & 1.13E+01           & 1.10E+01           & 1.11E+01           & 2.64E-01          & 1.57E-04 \\
            & GWO & 4.09E-08           & 1.10E-06           & 1.97E-07           & 6.64E-08           & 3.26E-07          & 1.57E-04 \\
            & IGA & \textbf{1.80E-10}  & \textbf{2.15E-09}  & \textbf{9.54E-10}  & \textbf{8.40E-10}  & \textbf{5.44E-10} & 1.00E+00 \\
        F11 & SGA & 1.55E+02           & 2.86E+02           & 2.31E+02           & 2.39E+02           & 4.51E+01          & 1.57E-04 \\
            & PSO & 1.13E+00           & 1.15E+00           & 1.14E+00           & 1.14E+00           & 6.09E-03          & 1.57E-04 \\
            & GWO & 1.23E-13           & 5.48E-01           & 5.48E-02           & 6.39E-13           & 1.73E-01          & 1.57E-04 \\
            & IGA & \textbf{0.00E+00}  & \textbf{5.55E-16}  & \textbf{3.11E-16}  & \textbf{3.89E-16}  & \textbf{2.15E-16} & 1.00E+00 \\ \bottomrule
        \end{tabular}
}
\end{table}

\begin{table}
\centering
\caption{Results of test functions (F1-F11) with 100 dimensions}
\label{tab:res3}
\resizebox{\textwidth}{!}{
        \begin{tabular}{@{}llllllll@{}}
        \toprule
        Fun & Alg & Min                & Max                & Mean               & Median             & Std               & P-value           \\ \midrule
        F1  & SGA & 7.51E+04           & 1.04E+05           & 8.49E+04           & 8.17E+04           & 8.94E+03          & 1.57E-04          \\
            & PSO & 1.26E+03           & 1.34E+03           & 1.29E+03           & 1.28E+03           & 3.11E+01          & 1.57E-04          \\
            & GWO & 6.89E-14           & 1.22E-11           & 2.26E-12           & 1.38E-12           & 3.61E-12          & 1.57E-04          \\
            & IGA & \textbf{6.51E-18}  & \textbf{1.87E-15}  & \textbf{4.94E-16}  & \textbf{2.71E-16}  & \textbf{5.43E-16} & 1.00E+00          \\
        F2  & SGA & 1.41E+33           & 1.50E+44           & 2.45E+43           & 1.33E+37           & 5.33E+43          & 1.57E-04          \\
            & PSO & 2.11E+31           & 8.05E+36           & 8.40E+35           & 2.90E+33           & 2.54E+36          & 1.57E-04          \\
            & GWO & 1.07E-10           & 4.50E-09           & 1.47E-09           & 9.56E-10           & 1.26E-09          & 2.12E-04          \\
            & IGA & \textbf{2.57E-12}  & \textbf{2.57E-10}  & \textbf{4.09E-11}  & \textbf{1.61E-11}  & \textbf{7.70E-11} & 1.00E+00          \\
        F3  & SGA & 3.37E+05           & 6.33E+05           & 4.57E+05           & 4.34E+05           & 8.89E+04          & 1.57E-04          \\
            & PSO & \textbf{1.33E+03}  & \textbf{2.03E+03}  & \textbf{1.67E+03}  & \textbf{1.68E+03}  & \textbf{2.31E+02} & 1.57E-04          \\
            & GWO & 7.96E+04           & 3.53E+05           & 1.93E+05           & 1.76E+05           & 8.50E+04          & \textbf{8.80E-01} \\
            & IGA & 1.04E+05           & 2.97E+05           & 1.88E+05           & 1.85E+05           & 6.23E+04          & 1.00E+00          \\
        F4  & SGA & 7.91E+01           & 9.15E+01           & 8.62E+01           & 8.61E+01           & 3.53E+00          & 1.57E-04          \\
            & PSO & 5.00E+00           & 5.00E+00           & 5.00E+00           & 5.00E+00           & 0.00E+00          & 1.57E-04          \\
            & GWO & 4.59E-02           & 9.09E+00           & 1.62E+00           & 3.06E-01           & 2.93E+00          & 1.57E-04          \\
            & IGA & \textbf{3.78E-06}  & \textbf{1.28E-03}  & \textbf{5.31E-04}  & \textbf{4.78E-04}  & \textbf{4.14E-04} & 1.00E+00          \\
        F5  & SGA & 1.67E+08           & 2.20E+08           & 1.93E+08           & 1.98E+08           & 2.09E+07          & 1.57E-04          \\
            & PSO & 2.39E+06           & 2.79E+06           & 2.58E+06           & 2.59E+06           & 1.38E+05          & 1.57E-04          \\
            & GWO & 9.83E+01           & 9.86E+01           & 9.84E+01           & 9.84E+01           & \textbf{8.77E-02} & 6.70E-04          \\
            & IGA & \textbf{9.82E+01}  & \textbf{9.83E+01}  & \textbf{9.83E+01}  & \textbf{9.83E+01}  & 5.72E-02          & 1.00E+00          \\
        F6  & SGA & 7.30E+04           & 1.01E+05           & 9.16E+04           & 9.19E+04           & 8.49E+03          & 1.57E-04          \\
            & PSO & 1.25E+03           & 1.31E+03           & 1.27E+03           & 1.27E+03           & 1.91E+01          & 1.57E-04          \\
            & GWO & 8.02E+00           & 1.23E+01           & 9.71E+00           & 9.63E+00           & 1.37E+00          & 1.57E-04          \\
            & IGA & \textbf{4.71E+00}  & \textbf{7.34E+00}  & \textbf{5.88E+00}  & \textbf{5.71E+00}  & \textbf{8.28E-01} & 1.00E+00          \\
        F7  & SGA & 3.22E+10           & 4.54E+10           & 3.97E+10           & 4.00E+10           & 4.34E+09          & 1.57E-04          \\
            & PSO & 1.15E+06           & 1.35E+06           & 1.22E+06           & 1.21E+06           & 6.48E+04          & 1.57E-04          \\
            & GWO & 1.75E-03           & 1.01E-01           & 3.96E-02           & 2.07E-02           & 4.05E-02          & 1.94E-03          \\
            & IGA & \textbf{1.03E-03}  & \textbf{6.57E-03}  & \textbf{2.61E-03}  & \textbf{2.10E-03}  & \textbf{1.71E-03} & 1.00E+00          \\
        F8  & SGA & -2.18E+04          & -1.77E+04          & -2.04E+04          & -2.07E+04          & 1.16E+03          & 1.94E-03          \\
            & PSO & -2.44E+02          & -1.89E+02          & -2.19E+02          & -2.20E+02          & \textbf{1.56E+01} & 1.57E-04          \\
            & GWO & -2.99E+04          & \textbf{-2.73E+04} & \textbf{-2.88E+04} & -2.91E+04          & 8.66E+02          & 1.91E-02          \\
            & IGA & \textbf{-3.02E+04} & -1.85E+04          & -2.87E+04          & \textbf{-2.99E+04} & 3.60E+03          & 1.00E+00          \\
        F9  & SGA & 6.22E+63           & 8.66E+66           & 1.88E+66           & 4.14E+65           & 2.81E+66          & 1.57E-04          \\
            & PSO & 1.91E+71           & 2.46E+71           & 2.20E+71           & 2.17E+71           & 1.52E+70          & 1.57E-04          \\
            & GWO & 1.14E-13           & 9.80E-09           & 1.08E-09           & 1.82E-11           & 3.07E-09          & 5.07E-04          \\
            & IGA & \textbf{0.00E+00}  & \textbf{2.27E-13}  & \textbf{9.09E-14}  & \textbf{0.00E+00}  & \textbf{1.17E-13} & 1.00E+00          \\
        F10 & SGA & 1.75E+01           & 1.91E+01           & 1.83E+01           & 1.85E+01           & 4.99E-01          & 1.57E-04          \\
            & PSO & 1.12E+01           & 1.17E+01           & 1.14E+01           & 1.14E+01           & 1.55E-01          & 1.57E-04          \\
            & GWO & 2.57E-08           & 2.07E-07           & 1.15E-07           & 1.14E-07           & 6.21E-08          & 1.57E-04          \\
            & IGA & \textbf{4.16E-10}  & \textbf{4.01E-09}  & \textbf{1.52E-09}  & \textbf{1.32E-09}  & \textbf{1.09E-09} & 1.00E+00          \\
        F11 & SGA & 6.60E+02           & 9.10E+02           & 7.63E+02           & 7.62E+02           & 7.79E+01          & 1.57E-04          \\
            & PSO & 1.30E+00           & 1.34E+00           & 1.33E+00           & 1.33E+00           & 1.02E-02          & 1.57E-04          \\
            & GWO & 1.80E-13           & 2.99E-11           & 4.23E-12           & 1.08E-12           & 9.08E-12          & 1.57E-04          \\
            & IGA & \textbf{1.11E-16}  & \textbf{1.22E-15}  & \textbf{4.44E-16}  & \textbf{3.89E-16}  & \textbf{3.70E-16} & 1.00E+00          \\ \bottomrule
        \end{tabular}
}
\end{table}

\begin{table}[]
        \centering
        \caption{Results of test functions (F12-15) with fixed dimensions}
        \label{tab:res4}
        \resizebox{\textwidth}{!}{
        \begin{tabular}{llllllllll}
                \toprule
                Fun & Alg & Min & Max & Mean & Median & Std & P-value \\
                \midrule
                F12 & SGA & 2.54E-04 & 6.31E-02 & 1.95E-02 & 6.19E-03 & 2.41E-02 & 1.57E-04 \\
                & PSO & 1.89E-07 & 3.39E-05 & 1.26E-05 & 6.56E-06 & 1.33E-05 & 1.57E-04 \\
                & GWO & 6.36E-18 & 1.97E-03 & 2.35E-04 & \textbf{4.81E-11} & 6.21E-04 & \textbf{5.45E-01} \\
                & IGA & \textbf{0.00E+00} & \textbf{9.31E-08} & \textbf{6.52E-08} & 9.31E-08 & \textbf{4.50E-08} & 1.00E+00 \\
                F13 & SGA & 4.34E-02 & 7.16E-01 & 4.88E-01 & 5.31E-01 & 2.04E-01 & 1.57E-04 \\
                & PSO & 1.51E-03 & 1.37E-02 & 6.55E-03 & 6.44E-03 & 3.92E-03 & 1.57E-04 \\
                & GWO & 1.95E-07 & 2.17E-01 & 2.86E-02 & \textbf{2.16E-04} & 6.81E-02 & \textbf{1.31E-01} \\
                & IGA & \textbf{4.44E-16} & \textbf{8.66E-04} & \textbf{4.33E-04} & 4.33E-04 & \textbf{4.56E-04} & 1.00E+00 \\
                F14 & SGA & 8.29E-04 & 1.54E-01 & 4.80E-02 & 2.40E-02 & 5.81E-02 & 2.85E-04 \\
                & PSO & \textbf{2.14E-06} & \textbf{1.29E-04} & \textbf{4.82E-05} & \textbf{3.86E-05} & \textbf{3.95E-05} & 3.20E-03 \\
                & GWO & 2.32E-01 & 5.95E+00 & 1.74E+00 & 1.13E+00 & 1.74E+00 & 1.57E-04 \\
                & IGA & 2.50E-05 & 1.44E-03 & 4.90E-04 & 3.25E-04 & 4.98E-04 & 1.00E+00 \\
                F15 & SGA & -9.54E+02 & -9.02E+02 & -9.39E+02 & -9.45E+02 & 1.64E+01 & 1.57E-04 \\
                & PSO & \textbf{-9.60E+02} & -5.72E+02 & -8.27E+02 & -8.81E+02 & 1.43E+02 & 2.50E-03 \\
                & GWO & \textbf{-9.60E+02} & -7.48E+02 & -8.65E+02 & -8.73E+02 & 8.05E+01 & 2.50E-03 \\
                & IGA & \textbf{-9.60E+02} & \textbf{-9.60E+02} & \textbf{-9.60E+02} & \textbf{-9.60E+02} & \textbf{7.05E-03} & 1.00E+00 \\
                \bottomrule
        \end{tabular}
        }
\end{table}

\newpage
\section{Application in Neural Network Adversarial Attack}
\subsection{MNST Dataset}
The MNST dataset (Mixed National Institute of Standards and Technology database)\citep{lecun1998mnist} is one of the most well-known datasets in the field of machine learning and is used in applications from simple experiments to published paper research.
It consists of handwritten digital images from 0-9. The MNIST image data is a single-channel grayscale map of $28\times28$ pixels, with each pixel taking values between 0 and 255, with 60,000 samples in the training set and 10,000 samples in the test set.
The general usage of the MNIST dataset is to learn with the training set first and then use the learned model to measure how well the test set can be correctly classified \citep{saito2018deep}.

\subsection{Implementation}
\begin{figure}[ht]
        \centering
        \subfigure[\label{subfig5-1}The structure of DCNN for experiment]{
                \includegraphics[width=0.55\textwidth]{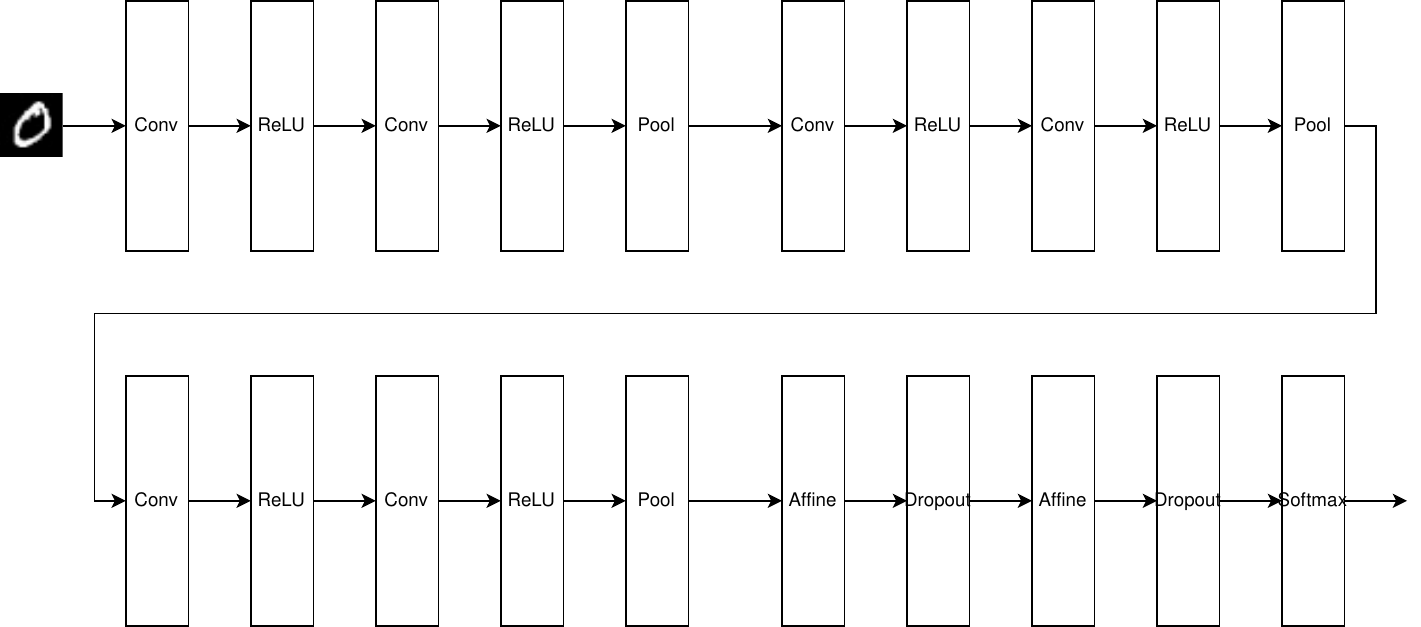}
        }
        \subfigure[\label{subfig5-2}The model of network adversarial attack]{
                \includegraphics[width=0.35\textwidth]{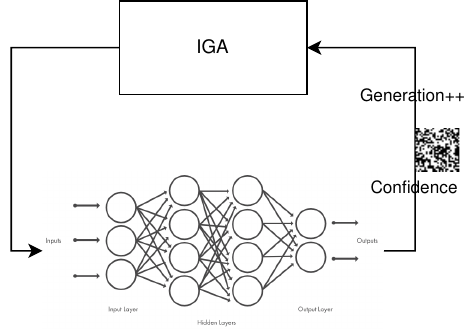}
        }
        \caption{The model of network adversarial attack}
\end{figure}

As shown in \figurename~\ref{subfig5-1}, the Deep Convolutional Neural Network (DCNN) pre-trained on the MNST dataset \citep{lecun1998mnist} is used as the experimental object in this paper, and the accuracy of the model is $99.35\%$ with a Loss value of 0.9632.
As shown in \figurename~\ref{subfig5-2}, the model of network adversarial attack is shown.
The number of populations of a specific size (set to 100 in this paper) is first generated and then input to the neural network to obtain the confidence of the specified labels.
To reduce the computational expense, the input is reduced to a binary image of $28\times28$ and the randomly generated binary image is iterated using the IGA proposed in this paper.
Among the 100 individuals, the fathers and mothers with relatively high confidence are selected by roulette selection, and then the children are generated by using the improved crossover link in this paper, and the children from a new population by improving the mutation link until the specified number of iterations.
Finally, the individual with the highest confidence is picked from the 100 individuals, which is the binary image with the highest confidence after passing through the neural network.

\subsection{Result}
\begin{figure}[ht]
        \centering
        \includegraphics[width=0.7\textwidth]{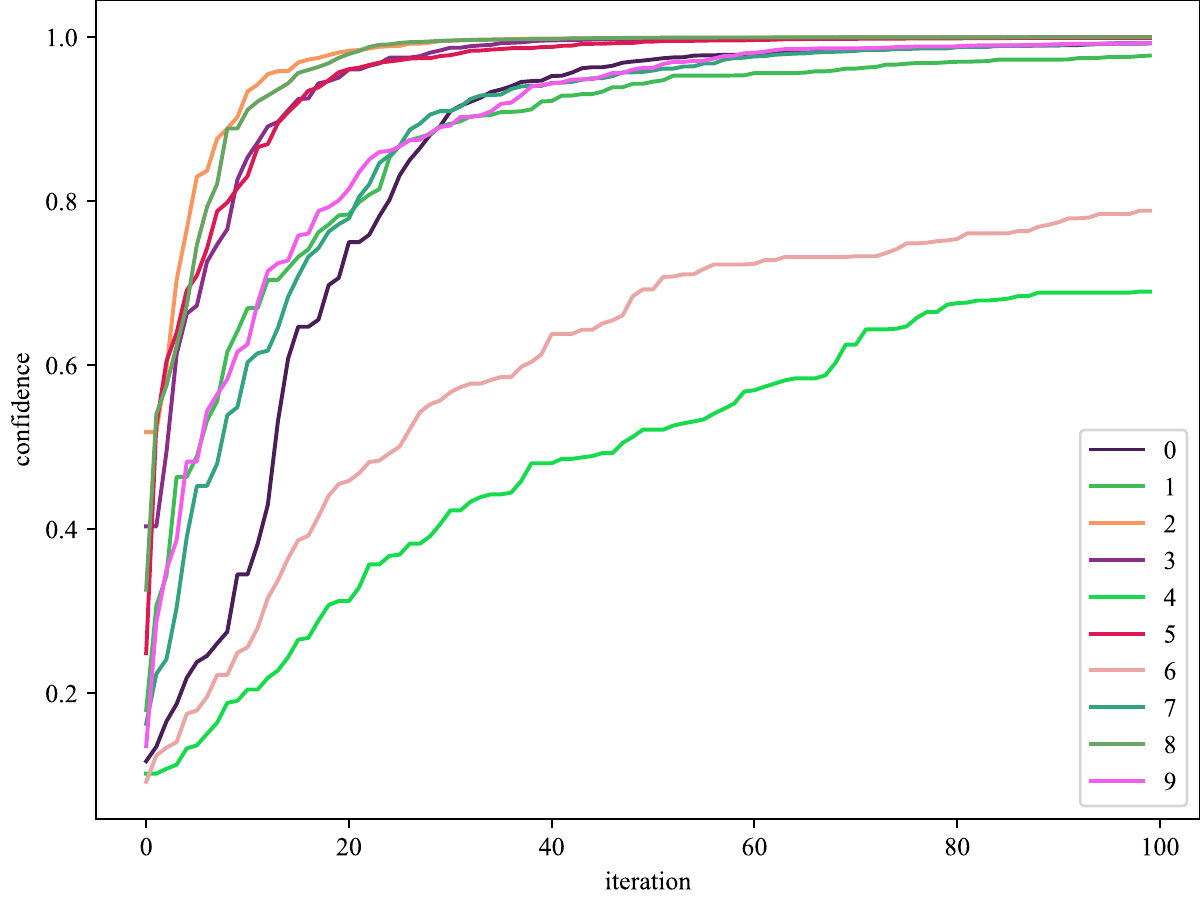}
        \caption{The confidence change of the binary image after iteration}
        \label{fig:cnn_result}
\end{figure}

As shown in \figurename~\ref{fig:cnn_result}, the confidence after 99 iterations of DCNN is $99.98\%$ for sample "2".
Sample "6" and sample "4" have the slowest convergence speed, and the confidence of sample "6" is $78.84\%$ after 99 iterations, and the confidence of sample "4" is $78.84\%$ after 99 iterations. 

The statistics of the experimental results are shown in \tablename~\ref{tab4}.
The binary image of sample "1" generated after 999 iterations has confidence of $99.94\%$ after passing DCNN, which is much higher than the confidence of sample "1" in the MNIST test set in the DCNN control group.
In the statistics of the results after initializing the population with the MNIST test set, because the overall confidence of the population initialized with the test set is higher, the increase in confidence during iteration is smaller.
The confidence of the sample selected from the MNIST test set is $99.56\%$, and after 10 iterations the confidence of the sample is $99.80\%$, and the number "1" becomes vertical; after 89 iterations the confidence is $99.98\%$, and the number "1 " has a tendency to "decompose" gradually.
\begin{table}[ht]
        \centering
        \caption{Statistical table of experimental results}
        \label{tab4}
        \begin{tabular}{ccccc}
                \toprule
                \figurename~& Label & Initialize & Iteration & Confidence \\
                \midrule
                \raisebox{-.5\height}{\includegraphics{result/1.jpg}} & 0 & random & - & $96.71\%$ \\
                \raisebox{-.5\height}{\includegraphics{result/2.jpg}} & 0 & random & 204 & $99.61\%$ \\
                \raisebox{-.5\height}{\includegraphics{result/3.jpg}} & 1 & random & - & $99.44\%$ \\
                \raisebox{-.5\height}{\includegraphics{result/4.jpg}} & 1 & random & 999 & $99.94\%$ \\
                \raisebox{-.5\height}{\includegraphics{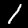}} & 1 & test dataset & - & $99.56\%$ \\
                \raisebox{-.5\height}{\includegraphics{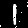}} & 1 & test dataset & 10 & $99.80\%$ \\
                \raisebox{-.5\height}{\includegraphics{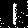}} & 1 & test dataset & 89 & $99.98\%$ \\
                \bottomrule
        \end{tabular}
\end{table}

\begin{figure}[!htp]
        \centering
        \includegraphics[width=0.7\textwidth]{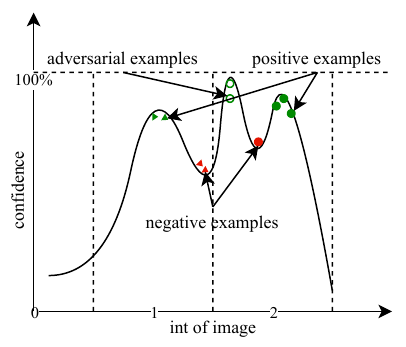}
        \caption{The confidence curve of a binary image}
        \label{fig:1drawio}
\end{figure}

As shown in \figurename~\ref{fig:1drawio}, the reason for this situation is probably that the confidence as a function of the image input is a multi-peak function, and the interval in which the test set images are distributed is not the highest peak of the confidence function.
This causes the initial population of the test set to "stray" from some pixels in the images generated by the IGA.

\newpage
\section{Conclusion}
The comparison and simulation experiments show that the improved method proposed in this paper is effective and greatly improves the convergence efficiency, global search capability and convergence success rate.
Applying IGA to the field of neural network adversarial attacks can also quickly obtain adversarial samples with high confidence, which is meaningful for the improvement of the robustness and security of neural network models.

In this paper, although the genetic algorithm has been improved to enhance the performance of the genetic algorithm, it is based on the genetic algorithm, so it cannot be completely separated from the general framework of the genetic algorithm, and the problem that the genetic algorithm is relatively slow in a single iteration cannot be solved.
We hope to explore a new nature-inspired optimization algorithm in our future work.
In addition, the reason why the neural network model has so many adversarial samples, we believe that it is a design flaw in the architecture of the neural network model.
In future work, we will also try to explore a completely new way of the infrastructure of neural networks so as to compress the space of adversarial samples.

With the wide application of artificial intelligence and deep learning in the field of computer vision, face recognition has outstanding performance in access control systems and payment systems, which require a fast response to the input face image, but this has instead become a drawback to be hacked.
For face recognition systems without in vivo detection, using the method in this paper only requires output labels and confidence information can obtain high confidence images quickly.
In summary, neural networks have many pitfalls due to their uninterpretability and still need to be considered carefully for use in important areas.

\newpage
\small

\bibliographystyle{apalike}
\bibliography{ecjsample}

\end{document}